%% file: sample-sigconf-authordraft.tex
\documentclass[sigconf]{acmart}
\renewcommand\footnotetextcopyrightpermission[1]{}
\usepackage{array}
\usepackage{lipsum}
\usepackage{colortbl}
\usepackage{textcomp}
\usepackage{stfloats}
\usepackage{url}
\usepackage{subcaption}
\usepackage{verbatim}
\usepackage{graphicx}
\usepackage{booktabs}
\usepackage{xcolor}
\usepackage{multirow}
\usepackage{multicol}
\usepackage{pifont}
\AtBeginDocument{%
  }

\newcommand{\eg}{\emph{e.g.}, }

\acmSubmissionID{338}

\begin{document}

\title{$\text{PKS}^4$:Parallel Kinematic Selective State Space Scanners\\ for Efficient Video Understanding}


\author{Lingjie Zeng}
\authornote{Lingjie Zeng and Hailun Zhang contributed equally to this work.}
\orcid{0009-0004-5546-6833}
\email{zenglingjie@stu.scu.edu.cn}
\affiliation{%
  \institution{College of Computer Science, Sichuan University}
  \city{Chengdu}
  \country{China}
}

\author{Hailun Zhang}
\authornotemark[1]
\orcid{0009-0002-5993-0416}
\email{tamakoko@stu.scu.edu.cn}
\affiliation{%
  \institution{College of Computer Science, Sichuan University}
  \city{Chengdu}
  \country{China}
}

\author{Xiwen Wang}
\affiliation{%
  \institution{College of Computer Science, Sichuan University}
  \city{Chengdu}
  \country{China}
}

\author{Qijun Zhao}
\authornote{Corresponding author.}
\orcid{0000-0003-4651-7163}
\email{qjzhao@scu.edu.cn}
\affiliation{%
  \institution{College of Computer Science, Sichuan University}
  \city{Chengdu}
  \country{China}
}

\renewcommand{\shortauthors}{Zeng et al.}



\input{Sections/0-abstract}


\begin{teaserfigure}
\centering
      \includegraphics[width=\textwidth]{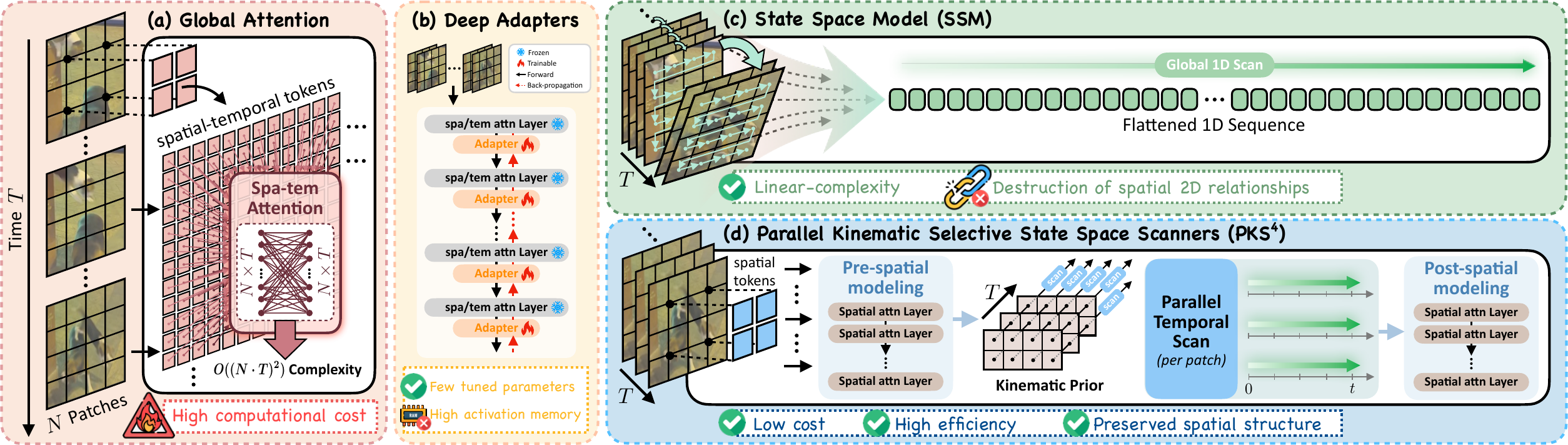}
  \caption{
(a) \emph{Global attention} suffers from quadratic computational complexity, leading to a massive computational bottleneck.
(b) \emph{Deep adapters} require storing intermediate activations across the entire backbone for back-propagation, severely suffering from an activation memory (VRAM) bottleneck.
(c) \emph{State Space Models (SSMs)} flatten 3D video tokens into a 1D sequence, destroying the innate 2D spatial relationships and consequently requiring exorbitant from-scratch pre-training.
(d) \emph{Our $\text{PKS}^4$} decouples the spatial-temporal modeling by leveraging a 2D vision backbone for spatial semantics and inserting parallel 1D temporal scanners, maintaining highly efficient and preserving spatial structures.
}
  \label{fig:teaser}
\end{teaserfigure}


\maketitle
\input{Sections/1-introduction}
\input{Sections/2-related_work}
\input{Sections/3-method}
\input{Sections/4-experiments}
\input{Sections/5-conclusion}

\bibliographystyle{ACM-Reference-Format}
\bibliography{sample-base}
\end{document}

%% file: Sections/0-abstract.tex
\begin{abstract}
Temporal modeling remains a fundamental challenge in video understanding, particularly as sequence lengths scale. 
Traditional video models relying on dense spatiotemporal attention suffer from quadratic computational costs for long videos.
To circumvent these costs, recent approaches adapt image models for videos via Parameter-Efficient Fine-Tuning (PEFT) methods such as adapters.
However, deeply inserting these modules incurs prohibitive activation memory overhead during back-propagation.
While recent efficient State Space Models (SSMs) introduce linear complexity, they disrupt 2D spatial relationships and consequently rely on extensive masked pre-training to recover spatial awareness.

To overcome these limitations, we propose \emph{\textbf{P}arallel \textbf{K}inematic \textbf{S}elective \textbf{S}tate \textbf{S}pace \textbf{S}canners} ($\text{PKS}^4$). 
We retain a standard 2D vision backbone for spatial semantics and insert a single plug-and-play $\text{PKS}^4$ module with linear-complexity temporal scanning, bypassing the need for temporal attention or multi-layer adapters.
We first explicitly extract kinematic priors via a Kinematic Prior Encoder, which captures local displacements and motion boundaries through inter-frame correlations and differences.
These priors subsequently drive linear-complexity SSMs to track the underlying kinematic states, adaptively modulating the update speeds and read-write strategies based on the input content at each time step.

Instead of global scanning, we deploy parallel scanners along the temporal dimension for each spatial location, preserving spatial structures while minimizing computational overhead.
Extensive experiments on spatial-heavy and temporal-heavy action recognition benchmarks demonstrate that $\text{PKS}^4$ achieves state-of-the-art performance. 
Remarkably, our method converges in merely $20$ epochs, achieving approximately $10\times$ lower training compute compared to pure video SSMs, thereby establishing a new paradigm for efficient video understanding.
\end{abstract}

%% file: Sections/1-introduction.tex
\section{Introduction}
\label{sec:introduction}
\begin{figure}[t]
    \centering
    \includegraphics[width=0.24\textwidth]{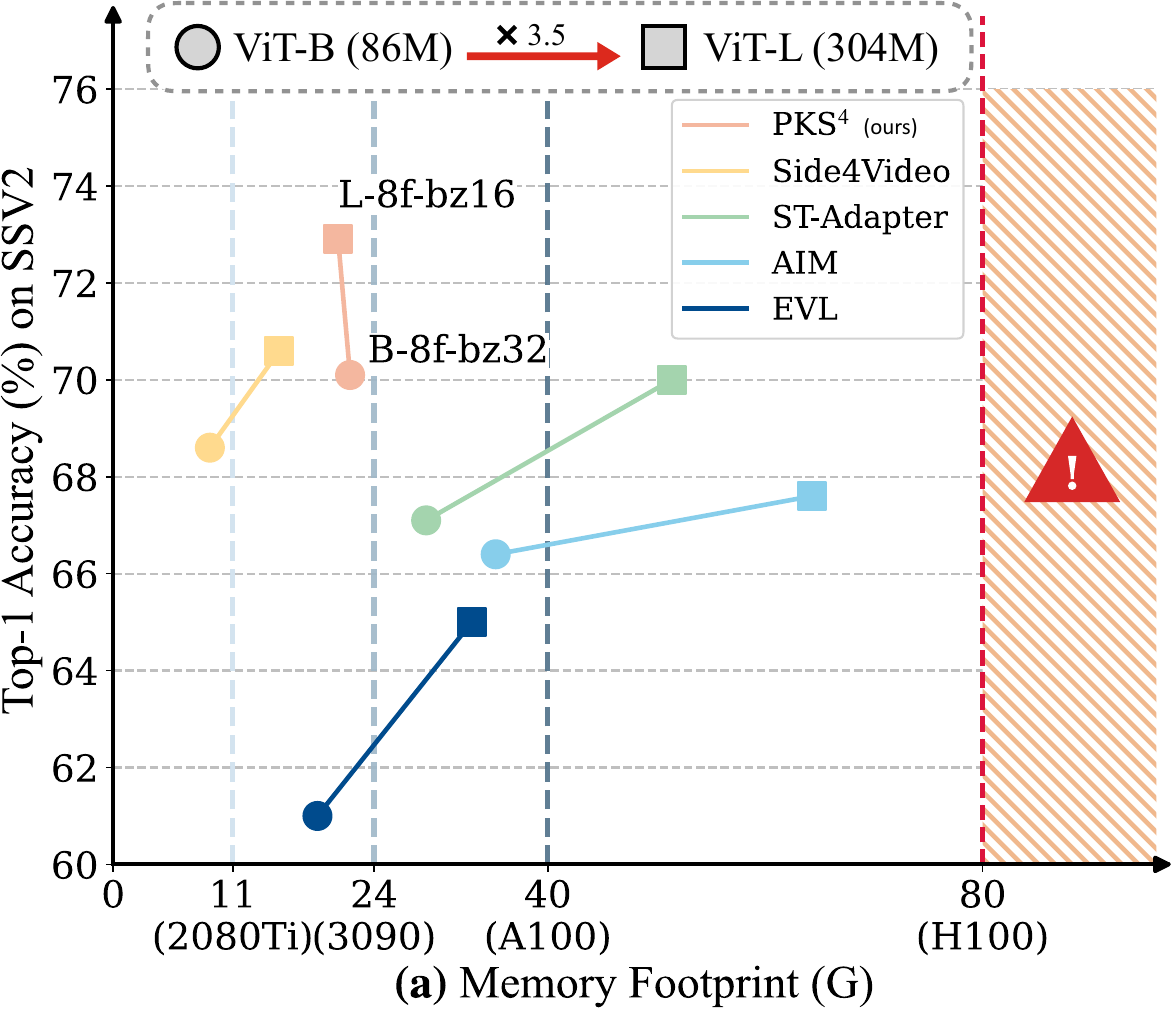}
    \hfill
    \includegraphics[width=0.23\textwidth]{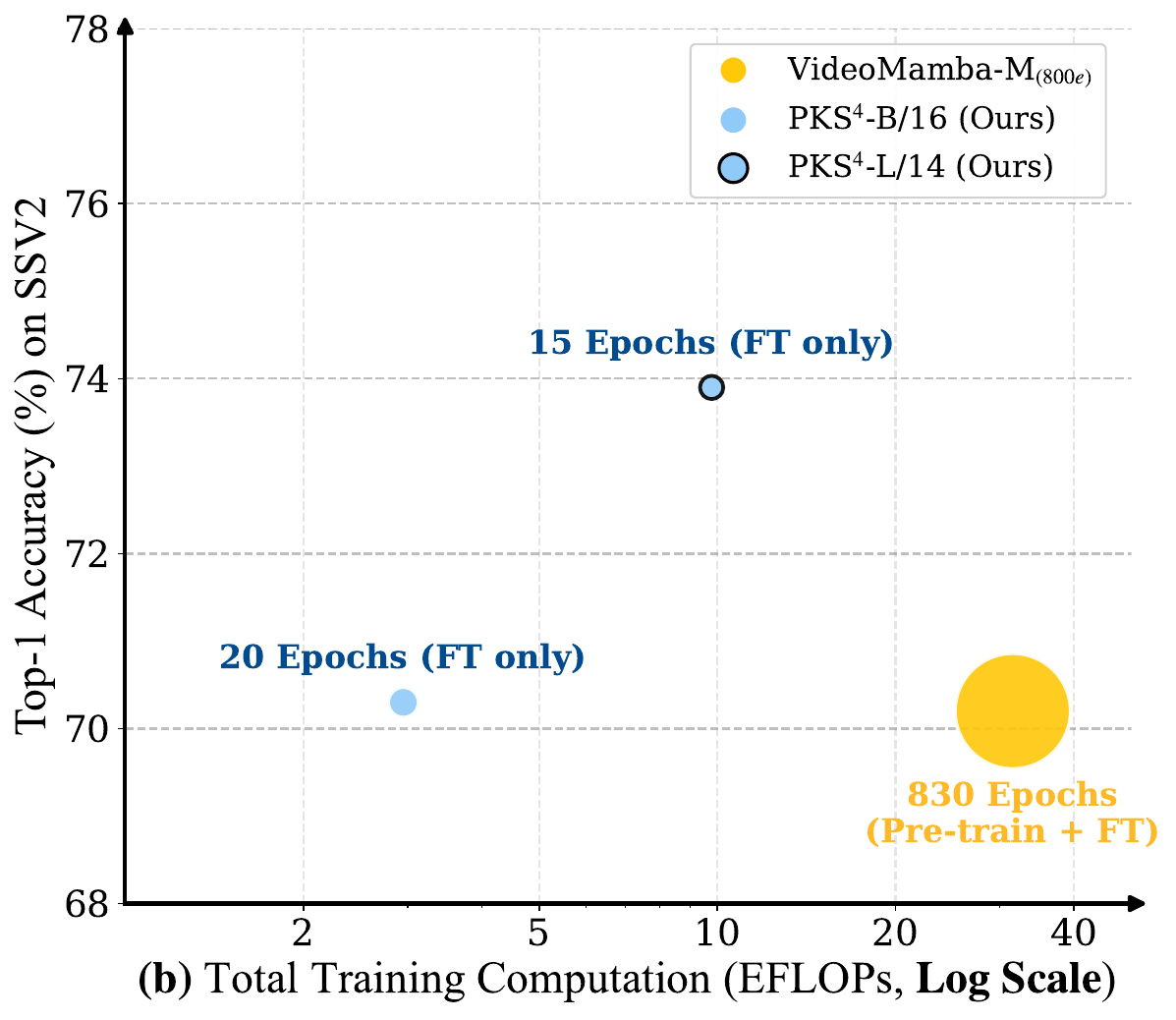}
    \caption{
    (a) Comparison of training memory usage with PEFT methods.
     (b) Comparison of total training epochs and computation with pure video SSM.
     }
    \label{fig:exp1}
\end{figure}
Video understanding requires modeling not only strong spatial semantics within individual frames but also long-range temporal dependencies across frames. 
This requirement becomes increasingly challenging as video models scale to longer clips and denser token representations.
A dominant line of work extends the Vision Transformer (ViT) from images to videos by performing dense spatiotemporal attention over all tokens~\cite{TimeSformer, ViViT}.
While effective, such designs inherit the quadratic complexity of global attention with respect to the spatiotemporal sequence length, rendering them expensive in terms of both computation and memory. 
Consequently, temporal modeling remains a central bottleneck in video understanding, particularly when transferring large image-pretrained backbones to the video domain under practical training budgets.
To alleviate this issue, another line of research adapts image backbones to videos through Parameter-Efficient Fine-Tuning (PEFT), such as inserting lightweight temporal adapters into pretrained Vision Transformers~\cite{ST-Adapter, AIM}. 
However, although adapter-based methods reduce the number of trainable parameters, deeply inserting these adapters requires storing massive intermediate activations across the entire backbone during back-propagation, which results in substantial activation memory overhead~\cite{Side4Video}. 
As illustrated in Fig.~\ref{fig:exp1}(a), this practical memory burden limits the accuracy-memory trade-off of adapter-based transfer.
More recently, State Space Models (SSMs), particularly selective SSMs such as Mamba~\cite{Mamba}, have emerged as an attractive alternative owing to their linear complexity. 
By formulating continuous-time system dynamics into discrete state updates, SSMs introduce a linear-complexity paradigm with respect to sequence length, providing a solution to the quadratic bottleneck of global attention~\cite{S4}. 
This advantage has motivated a growing effort to apply SSMs to video tasks~\cite{VideoMamba}. 

Despite the success of SSMs in natural language processing, the naive application of pure SSMs to video understanding leaves two critical issues unresolved.
\textbf{(i)} As shown in Fig~\ref{fig:teaser}(c), videos are inherently 3D structures comprising spatial and temporal dimensions, a common strategy is to flatten the video into a 1D token sequence and subsequently perform continuous scanning. 
Although computationally efficient, this treatment mixes spatial and temporal ordering into a single sequence, which has been noted to be challenging due to the difficulty in jointly modeling temporal causality and spatial information~\cite {Vivim}.
Because temporal dependencies are intrinsically tied to spatial variations (\eg, object shifts, boundary changes, and local detail transformations), such destruction of the spatial structure leads to a lack of explicit understanding of \emph{what} changes and \emph{where} these changes originate.
To recover this lost spatial awareness, pure video SSMs are forced to rely on extensive from-scratch pre-training, which is computationally prohibitive and inefficient for bridging the gap between image and video modalities. 
As shown in Fig.~\ref{fig:exp1}(b), this reliance translates into substantially higher end-to-end training compute.
\textbf{(ii)} Furthermore, the content driving state transitions is often generic token content rather than motion-aware kinematic state content, which is essential for video understanding.
These issues highlight the demand for an SSM-integrated video architecture that is not only scan-efficient but also capable of determining the appropriate content to drive state transitions, thereby preserving alignment with the kinematic nature of videos.
To this end, we propose \emph{\textbf{P}arallel \textbf{K}inematic \textbf{S}elective \textbf{S}tate \textbf{S}pace \textbf{S}canners} ($\text{PKS}^4$).
Instead of replacing the spatial backbone or densely injecting temporal adapters across multiple layers, we retain a standard 2D image-pretrained ViT as the primary backbone for spatial representation learning and insert a \emph{single} plug-and-play temporal module at an intermediate layer. 
Before performing selective state-space scanning, we explicitly enrich the input content with motion-aware kinematic priors. 
Concretely, given intermediate tokens from a ViT block, we reshape the patch tokens back to their spatial layout and process them through a \textbf{Kinematic Prior Encoder}, as illustrated in Fig~\ref{fig:teaser}(d). 
This encoder applies a correlation operator and a variation operator sequentially to capture two complementary forms of motion patterns: 
(i) inter-frame correspondences reflecting local displacement correlations, and (ii) inter-frame changes highlighting motion boundaries and temporal variations. 
Such explicit kinematic priors provide a more appropriate state context for SSMs, encouraging the scanner to model motion-driven state transitions rather than directly processing raw, flattened video tokens. 
Building upon these kinematic priors, $\text{PKS}^4$ performs temporal modeling via parallel patch-wise scanning. 
Instead of flattening all spatial and temporal tokens into a single 1D sequence, we construct an individual temporal sequence for each spatial location and scan these sequences in parallel along the temporal dimension. 
This design preserves the 2D spatial structure inherited from the image backbone while maintaining linear complexity for temporal modeling.  
The parallel scanning mechanism dynamically generates data-dependent parameters to adaptively modulate update speeds and read-write strategies based on the input content.
Following the scanning process, the updated tokens are reshaped and fed into the subsequent ViT layer, rendering the module \emph{fully compatible} with the original backbone. 
Consequently, $\text{PKS}^4$ achieves linear-complexity temporal tracking while preserving the innate spatial structure of the video, providing a simple, robust route for transferring strong image backbones to video understanding.
Extensive experiments on temporal-heavy and spatial-heavy benchmarks validate the effectiveness of this design. 
More importantly, these results support the broader perspective that efficient temporal modeling does not necessitate a strict choice between dense attention architectures and SSMs. 
Given an appropriate temporal interface, SSMs can be made substantially more compatible with the structure of the ViT.
The core contributions of this work are summarized as follows:

\begin{itemize}
    \item We propose Parallel Kinematic Selective State Space Scanners ($\text{PKS}^{4}$), a plug-and-play module that achieves linear-complexity temporal modeling by synergizing kinematic priors with SSM, avoiding the computational bottlenecks of spatiotemporal attention and deep adapters.
    \item We introduce a Kinematic Prior Encoder to explicitly extract motion-aware kinematic priors, providing an appropriate task-specific state context for SSMs.
    \item We decouple the spatial and temporal dimensions by deploying parallel scanners to preserve the spatial structure inherited from pre-trained backbones, effectively avoiding exorbitant from-scratch pre-training.
    \item Extensive experiments demonstrate that $\text{PKS}^{4}$ achieves state-of-the-art performance on spatial-heavy and temporal-heavy action recognition benchmarks while keep efficient.
    Remarkably, our method converges in merely $20$ epochs, requiring approximately $10\times$ lower training compute than pure video SSMs, establishing a new paradigm for highly efficient video understanding.
\end{itemize}

%% file: Sections/2-related_work.tex
\section{Related Work}
\label{sec:related_work}
\subsection{Spatial-Temporal Attention}
A mainstream paradigm in video understanding extends Vision Transformers from images to videos by jointly computing attention over spatiotemporal tokens.
Early representative works~\cite{ViViT,TimeSformer} adopt this approach to process sequences of spatiotemporal tokens, achieving strong performance in video recognition. 
Subsequent efforts further improve the trade-off between efficiency and accuracy.
For example, MViT~\cite{MViT} constructs multiscale feature hierarchies for visual sequences. 
Additionally, Swin Transformer~\cite{SwinTransformer} constrains attention through local windows and shifting mechanisms, while UniFormer~\cite{UniFormer} combines convolution-like local aggregation with global token interaction to better balance locality and long-range dependencies.
However, these methods inherit the quadratic complexity of global attention with respect to the total number of video tokens, leading to substantial computational overhead as the clip length or the spatial resolution increases.
Unlike prior works that directly entangle spatial and temporal modeling through global attention, the proposed method decouples these two processes. 
We retain a standard 2D image-pretrained ViT to model spatial semantics, while employing a lightweight plug-and-play module for linear-complexity temporal modeling.

\subsection{Parameter-Efficient Temporal Adapter}
Another line of research aims to adapt image-pretrained models to videos through parameter-efficient fine-tuning.
Specifically, ST-Adapter~\cite{ST-Adapter} introduces lightweight spatiotemporal adapters into frozen image transformers. 
Furthermore, AIM~\cite{AIM} incrementally equips pretrained image models with spatial, temporal, and joint adapters, and DualPath~\cite{DUALPATH} separates the spatial and temporal adaptation processes into two distinct lightweight streams.
However, existing work indicates that reducing the number of trainable parameters does not necessarily alleviate the actual training burden~\cite{Side4Video}. 
In many of these methods, temporal adaptation is distributed across multiple layers of the backbone network via repeated adapter insertions.
Consequently, back-propagation still requires storing a large volume of intermediate activations across the network.
In contrast, the proposed approach does not rely on the deep injection of multi-layer adapters.
Instead, we insert a single plug-and-play temporal module at an intermediate layer of the backbone, thereby avoiding the cumulative burden of activation memory.

\subsection{State Space Models}
State Space Models (SSMs) have recently emerged as an efficient alternative to Transformers due to their linear complexity with respect to the sequence length~\cite{Mamba,S4}. 
The work of Vim~\cite{Vim} demonstrates that bidirectional Mamba blocks can serve as an efficient visual backbone for images, thereby motivating subsequent attempts to extend SSMs to videos.
Building upon this advantage, VideoMamba~\cite{VideoMamba} directly adapts Mamba-style selective state transitions to achieve efficient video understanding, while the Video Mamba Suite~\cite{VideoMambaSuite} systematically investigates the multiple roles that Mamba can play across a broad set of video tasks. 
Despite their efficiency, existing SSM approaches for videos typically serialize the video tokens into a flattened one-dimensional sequence and subsequently apply global scanning.
This process destroys the native two-dimensional spatial structure of the video patches and results in the under-utilization of pre-trained image knowledge.
Consequently, the model must be trained from scratch to recover spatial awareness.
Furthermore, the content driving the state transitions is often generic token content, rather than the motion-aware kinematic content that is essential for video understanding. 
In this work, we address these issues by explicitly injecting motion-aware kinematic priors into the state transition process and performing parallel temporal scanning for each spatial location, thereby preserving the spatial structure inherited from the 2D backbone.

%% file: Sections/3-method.tex
\section{Method}
\label{sec:method}
We first introduce the overview of propose $\text{PKS}^4$ in \S~\ref{sec:method_overview}.
We then provide a detailed explanation of the spatial backbone, \emph{Kinematic Prioir Encdoer} (KPE), and \emph{Kinemetic Selecitve State Space Scanner} ($\text{KS}^4$) in \S~\ref{sec:method_backbone}, \S~\ref{sec:method_kpe}, and \S~\ref{sec:method_ks4}, respectively.
The training details are given in \S~\ref{sec:method_train_infer}.
\begin{figure*}
    \centering
    \includegraphics[width=\linewidth]{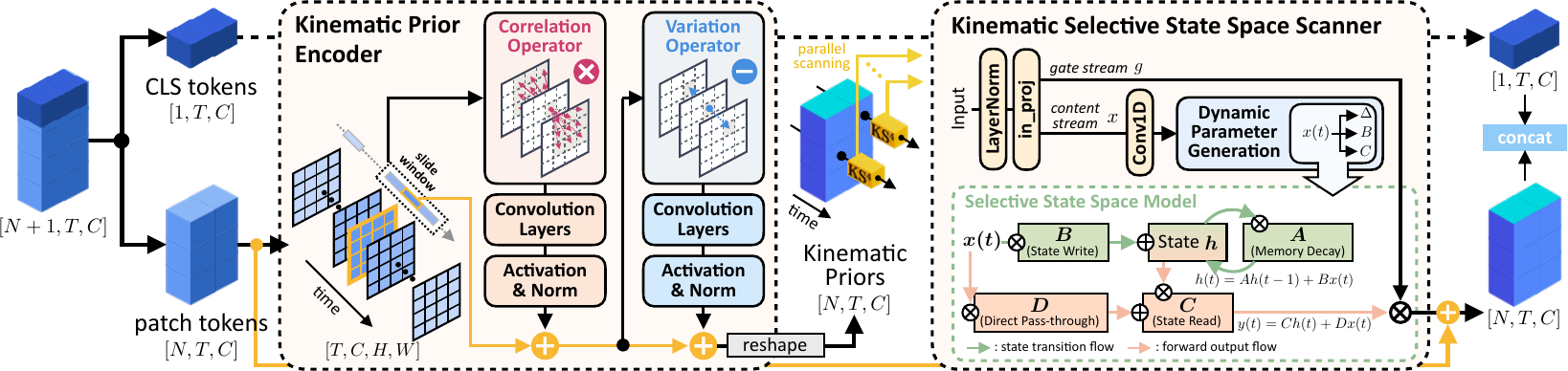}
    \caption{Overview of $\textbf{PKS}^4$.
    Given intermediate token features from a ViT layer, we first split CLS and patch tokens.
    The Kinematic Prior Encoder process patch tokens through a temporal sliding window.
    The correlation and variation operators are sequentially applied to explicitly capture rich kinematic priors.
    To preserve innate spatial structures, we deploy parallel scanning along the temporal dimension for each spatial location.
    Each Kinematic Selective State Space Scanner ($\textbf{KS}^4$) dynamically generates data-dependent parameters to adaptively modulate the update and decay processes, achieving linear-complexity temporal modeling.
    The updated tokens are concatenated and fed back to the next ViT layer.
    }
    \label{fig:method}
\end{figure*}

\subsection{Overview}
\label{sec:method_overview}
\noindent\textbf{Spatial encoding.}
Given an input video clip
$
\mathbf{X} \in \mathbb{R}^{B \times T \times 3 \times H \times W},
$
where $B$ denotes the batch size and $T$ denotes the number of frames, we first reshape the video into $B\!\cdot\!T$ images of resolution $H \times W$.
Then feed them into an image-pretrained ViT as the primary spatial encoder. 
For each frame, the backbone performs patch embedding and spatial token mixing as in the original image model. 
Let the intermediate features at the $l$-th layer be
$
\mathbf{Z}^{(l)} \in \mathbb{R}^{B \times T \times (N+1) \times C},
$
where $N$ is the number of spatial patch tokens and $C$ is the channel dimension. 
As shown in Fig.~\ref{fig:method}, we separate $\mathbf{Z}^{(l)}$ into a CLS token and patch tokens:
\begin{equation}
\mathbf{z}_{\mathrm{cls}}^{(l)} \in \mathbb{R}^{B \times T \times C}, \qquad
\mathbf{Z}_{\mathrm{p}}^{(l)} \in \mathbb{R}^{B \times T \times N \times C}.
\end{equation}
For spatial semantics extraction, we enhance the patch tokens with kinematic priors via the KPE module:
\begin{equation}
\widehat{\mathbf{Z}}_{\mathrm{p}}^{(l)} = \mathrm{KPE}\big(\mathbf{Z}_{\mathrm{p}}^{(l)}\big).
\end{equation}
 
\noindent\textbf{Parallel temporal scanning.}
Given the kinematic-prior-enhanced patch tokens $\widehat{\mathbf{Z}}_{\mathrm{p}}^{(l)}$, we form a temporal sequence for each spatial location, instead of flattening all $N \times T$ tokens into a single global sequence.
We then apply selective state space scanning only along the temporal axis.
For the $n$-th spatial location, we define a patch-wise temporal trajectory as:
\begin{equation}
\mathbf{s}_n = \left[
\widehat{\mathbf{Z}}_{\mathrm{p}}^{(l)}(:,1,n,:), \dots,
\widehat{\mathbf{Z}}_{\mathrm{p}}^{(l)}(:,T,n,:)
\right] \in \mathbb{R}^{B \times T \times C}.
\end{equation}
All trajectories are then processed in parallel with a shared scanner $\mathrm{KS}^4$:
\begin{equation}
\widetilde{\mathbf{s}}_n = \mathrm{KS}^4(\mathbf{s}_n), \qquad n=1,\dots,N.
\end{equation}
Finally, the scanned trajectories are reassembled according to their original spatial indices and added back to the pre-KPE patch tokens,
\begin{equation}
\widetilde{\mathbf{Z}}_{\mathrm{p}}^{(l)}
=
\mathbf{Z}_{\mathrm{p}}^{(l)}
+
\mathrm{Reassemble}\Big(
\{\widetilde{\mathbf{s}}_n\}_{n=1}^{N}
\Big)
\in \mathbb{R}^{B \times T \times N \times C}.
\label{eq:overview_pipeline}
\end{equation}

\noindent\textbf{Auxiliary CLS temporal route.}
In addition to the patch-wise scanner, we maintain a lightweight temporal route for the CLS token across backbone layers. 
Since the CLS token aggregates global semantics, a small temporal operator is sufficient to allow information exchange across frames. 
The CLS toekn is processed by a depthwise 1D convolution with kernel size $3$ along time:
\begin{equation}
\widetilde{\mathbf{z}}_{\mathrm{cls}}^{(l)} =
\mathrm{Conv1D}_{\mathrm{cls}}^{k=3}\!\left(\mathbf{z}_{\mathrm{cls}}^{(l)}\right),
\end{equation}
where the convolution is initialized in a shift-style manner, enabling weak temporal propagation from the start of training.
The updated patch tokens are then concatenated with the temporally enhanced CLS tokens and fed back to the next layer, without modifying the internal structure of the original ViT block. 
\begin{equation}
\widetilde{\mathbf{Z}}^{(l)} =
\mathrm{Concat}\!\left(
\widetilde{\mathbf{z}}_{\mathrm{cls}}^{(l)},
\widetilde{\mathbf{Z}}_{\mathrm{p}}^{(l)}
\right).
\end{equation}
\noindent\textbf{Video-level prediction.}
After the final ViT layer, we obtain frame-wise representations
$
\mathbf{v}_1,\dots,\mathbf{v}_T \in \mathbb{R}^{C}.
$
We aggregate them through temporal average pooling:
\begin{equation}
\mathbf{v}
=
\frac{1}{T}\sum_{t=1}^{T}\mathbf{v}_t,
\end{equation}
and feed the pooled feature into a linear classifier:
\begin{equation}
\label{eq:final_pred}
\mathbf{p} = \mathbf{W}_{\mathrm{cls}}\mathbf{v} + \mathbf{b}_{\mathrm{cls}}.
\end{equation}

\subsection{Spatial Backbone}
\label{sec:method_backbone}
We use an image-pretrained ViT as the spatial backbone.
Given the input video, we reshape it into $B\!\cdot\!T$ frames and tokenize each frame into $N$ patch tokens together with one CLS token.
After positional encoding, the tokens are processed by standard Transformer blocks.
Let
$
\mathbf{Z}^{(i-1)} \in \mathbb{R}^{B \times T \times (N+1) \times C}
$
be the input to the $i$-th block, the forward pass if formulated as:
\begin{equation}
\mathbf{Z}^{(i-\frac{1}{2})}
=
\mathbf{Z}^{(i-1)}
+
\mathrm{MSA}\big(\mathrm{LN}(\mathbf{Z}^{(i-1)})\big),
\end{equation}
\begin{equation}
\mathbf{Z}^{(i)}
=
\mathbf{Z}^{(i-\frac{1}{2})}
+
\mathrm{MLP}\big(\mathrm{LN}(\mathbf{Z}^{(i-\frac{1}{2})})\big),
\end{equation}
where MSA, MLP, and LN denote Multi-head Self-Attention~\cite{ViT}, Multi-Layer Perceptron~\cite{ViT}, and Layer Normalization~\cite{LN} respectively.
Our plug-and-play temporal module $\text{PKS}^4$ is inserted only once after the $l$-th block.

\subsection{Kinematic Prior Encoder}
\label{sec:method_kpe}
To provide compatible input with the kinematic nature of videos for the state transition, we introduce the \textbf{Kinematic Prior Encoder (KPE)} that explicitly extracts motion-aware priors from intermediate patch tokens before scanning.
Given patch tokens
$
\mathbf{Z}_{\mathrm{p}} \in \mathbb{R}^{B \times T \times N \times C},
$
we first reshape them back to a spatial grid:
\begin{equation}
\mathbf{F} \in \mathbb{R}^{B \times T \times C \times H_p \times W_p},
\qquad N = H_p W_p .
\end{equation}
The KPE processes the spatial feature $\mathbf{F}$ sequentially using two complementary operators:
\begin{equation}
\mathbf{F}^{\mathrm{corr}} = \mathcal{O}_{\mathrm{corr}}(\mathbf{F}), \qquad
\mathbf{F}^{\mathrm{var}}  = \mathcal{O}_{\mathrm{var}}(\mathbf{F}^{\mathrm{corr}}),
\end{equation}
where $\mathcal{O}_{\mathrm{corr}}$ captures inter-frame correspondence patterns and $\mathcal{O}_{\mathrm{var}}$ emphasizes temporal changes and motion boundaries. 
The resulting features are projected back to the token space and added to the original patch representation:
\begin{equation}
\widehat{\mathbf{Z}}_{\mathrm{p}}
=
\mathbf{Z}_{\mathrm{p}} + \phi\!\left(\mathbf{F}^{\mathrm{var}}\right),
\label{eq:kpe_output}
\end{equation}
where $\phi(\cdot)$ denotes reshaping and projection back to $\mathbb{R}^{B \times T \times N \times C}$.

\subsubsection{Correlation Operator}
\label{sec:corr_op}
The correlation operator is designed to model local displacement consistency across neighboring frames. 
For a target spatial location $p$ at time $t$, we compare the feature $\mathbf{f}_{t}(p) \in \mathbb{R}^{C}$ at $p$ with features within a local search window in nearby frames. 
For a temporal offset $\tau$ and spatial displacement $\delta = (u,v)$, we define the local correlation score as:
\begin{equation}
\mathcal{S}_{t,\tau}(p,\delta)
=
\left\langle
\bar{\mathbf{f}}_{t}(p),
\bar{\mathbf{f}}_{t+\tau}(p+\delta)
\right\rangle,
\label{eq:corr_basic}
\end{equation}
where $\bar{\mathbf{f}}$ denotes channel-wise normalized features and $\langle \cdot,\cdot \rangle$ is the inner product. 
The temporal offset $\tau$ is restricted to a local temporal window, and $\delta$ is restricted to a local spatial region.
In this way, the operator encodes whether a local pattern at time $t$ reappears nearby at time $t+\tau$, thereby extracting local motion and correspondence.
All correlation responses within the temporal-spatial neighborhood are then aggregated and mapped back to the feature domain:
\begin{equation}
\mathbf{F}^{\mathrm{corr}}
=
\psi_{\mathrm{corr}}
\big(
\{
\mathcal{S}_{t,\tau}(p,\delta)
\}_{\tau,\delta}
\big),
\end{equation}
where $\psi_{\mathrm{corr}}(\cdot)$ denotes the learnable projection to the visual feature space. 
Compared with global attention, this operator does not reweigh all tokens globally, but instead extracts displacement-aware evidence from local temporal neighborhoods.

\subsubsection{Variation Operator}
\label{sec:var_op}
While the correlation operator focuses on matching and displacement, the variation operator focuses on temporal change. 
It captures motion boundaries, appearance changes, and dynamic regions through inter-frame difference modeling.
Given the intermediate feature map, we compute temporal differences within a local temporal context. 
For a temporal offset $\tau$, the variation signal is defined as:
\begin{equation}
\Delta_{t,\tau}(p)
=
\mathbf{f}_{t+\tau}(p) - \mathbf{f}_{t}(p).
\label{eq:var_basic}
\end{equation}
When multiple neighboring frames are considered, we collect a set of difference responses $\{\Delta_{t,\tau}(p)\}_{\tau}$ and jointly aggregates them:
\begin{equation}
\mathbf{F}^{\mathrm{var}}
=
\psi_{\mathrm{var}}
\big(
\{
\Delta_{t,\tau}(p)
\}_{\tau}
\big),
\end{equation}
where $\psi_{\mathrm{var}}(\cdot)$ is a learnable projection.
By complementarily emphasizes \emph{where similar content moves} and \emph{where content changes}, KPE provides kinematic priors for the subsequent scanner. 

\subsection{Kinematic Selective State Space Scanner}
\label{sec:method_ks4}
After KPE, the patch tokens become motion-aware:
$
\widehat{\mathbf{Z}}_{\mathrm{p}} \in \mathbb{R}^{B \times T \times N \times C}.
$
We then apply the \textbf{Kinematic Selective State Space Scanner (KS$^4$)} to each spatial location independently along the temporal axis. 
Specifically, we reshape the tokens into $\widehat{\mathbf{Z}}_{\mathrm{seq}}\in\mathbb{R}^{(B N) \times T \times C}$ so that each row corresponds to one patch-wise temporal sequence.

\noindent\textbf{State parameterization.}
As shown in Fig~\ref{fig:method}, for a patch-wise temporal sequence $\mathbf{u} \in \mathbb{R}^{T \times C}$, we first apply normalization.
Then, a linear projection expands the channels and splits the result into a content stream and a gate stream:
\begin{equation}
[\mathbf{x}^{\prime}, \mathbf{g}]
=
\mathbf{W}_{\mathrm{in}} \mathrm{LN}(\mathbf{u}),
\end{equation}
where $\mathbf{x}$ is the content branch and $\mathbf{g}$ is the gating branch.
A local temporal mixing operator is then applied to the content embedding:
\begin{equation}
\mathbf{x} = \mathrm{Conv1D}\!\left(\mathbf{x}^{\prime}\right).
\end{equation}
Following selective SSMs, the dynamic parameters are generated from the input itself:
\begin{equation}
[\boldsymbol{\delta}, \mathbf{B}, \mathbf{C}]
=
\Gamma(\mathbf{x}),
\label{eq:dynamic_params}
\end{equation}
where $\Gamma(\cdot)$ denotes a learnable projection. 
Here, $\boldsymbol{\delta}$ controls the input-dependent step size, while $\mathbf{B}$ and $\mathbf{C}$ parameterize the input-dependent write and read operations. 
The step size is constrained to be positive via:
\begin{equation}
\Delta_t = \mathrm{softplus}(\delta_t + b_{\delta}).
\end{equation}

\noindent\textbf{Selective state transition.}
Let $\mathbf{h}_t \in \mathbb{R}^{d_s}$ denote the hidden state at time step $t$, where $d_s$ is the state dimension. 
The selective state transition follows the standard SSM form:
\begin{equation}
\mathbf{h}_t
=
\mathbf{A}(\Delta_t)\mathbf{h}_{t-1}
+
\mathbf{B}_t \mathbf{x}_t,
\label{eq:ssm_state}
\end{equation}
\begin{equation}
\mathbf{y}_t
=
\mathbf{C}_t \mathbf{h}_t
+
\mathbf{D}\mathbf{x}_t,
\label{eq:ssm_output}
\end{equation}
where $\mathbf{A}(\Delta_t)$ is the discretized transition matrix controlled by the step size $\Delta_t$, $\mathbf{B}_t$ and $\mathbf{C}_t$ are write/read parameters, and $\mathbf{D}$ is the direct skip term. 
The scanner is driven by motion-aware tokens produced by the KPE. 
Therefore, the data-dependent parameters in Eqns.~\eqref{eq:dynamic_params}--\eqref{eq:ssm_output} are not generated from generic flattened video tokens, but from kinematically enriched features that already encode local displacement and temporal variation. 
This makes the state evolution more compatible with video dynamics.
The final output of the scanner is obtained through output projection and residual fusion:
\begin{equation}
\widetilde{\mathbf{u}}_t
=
\mathbf{W}_{\mathrm{out}}
\big(
\mathbf{y}_t \odot \sigma(\mathbf{g}_t)
\big)
+
\mathbf{u}_t,
\end{equation}
where $\sigma(\cdot)$ is a gating nonlinearity.

\subsection{Training Details}
\label{sec:method_train_infer}
\subsubsection{Training Objectives}
We train the model using the standard cross-entropy loss for video classification. 
Given the predicted logits $\mathbf{p} \in \mathbb{R}^{C}$ over $C$ action classes and the ground-truth label distribution $\mathbf{q} \in \mathbb{R}^{C}$, the classification loss is defined as:
\begin{equation}
\mathcal{L}
= - \sum_{c=1}^{C} q_c \log \left( \frac{\exp(p_c)}{\sum_{j=1}^{C} \exp(p_j)} \right),
\end{equation}
where $q_c$ denotes the target probability of class $c$. 
With label smoothing, $\mathbf{q}$ is a smoothed version of the one-hot label distribution. 

\subsubsection{Optimization Strategy}
We use AdamW for optimization and adopt a cosine learning-rate schedule with warm-up. 
The learning rate is scaled according to the global batch size. 
To better preserve pre-trained representations while enabling efficient adaptation of the newly inserted $\text{PKS}^4$ modules, we apply layer-wise learning-rate decay to the ViT backbone. 
Newly introduced parameters are exempted from aggressive decay, allowing them to adapt faster to video dynamics.
Our implementation also applies stable initialization for newly inserted modules, including near-identity initialization in residual projections and conservative initialization for temporal operators. 
This reduces destructive interference with the pre-trained image backbone in early training and makes single-module insertion more effective.

%% file: Sections/4-experiments.tex
\section{Experiments}
\label{sec:experiments}
\subsection{Datasets and Evaluation Metrics}
We evaluate VolRF on two public action recognition benchmarks.
\textbf{(i) Something-Something V2} (SSV2)~\cite{ssv2} consists of $174$ fine-grained action classes that are object and motion-centric. 
It requires learning the subtle differences between motions and object interactions.
The training, validation, and test sets contain $168,913$, $24,777$, and $27,157$ videos, respectively. 
\textbf{(ii) Kinetics-400} (K400)~\cite{k400} covers about 240K training videos of $400$ categories and 20K test videos sourced from YouTube, consisting of appearance-balanced classes of everyday human actions derived from real-world scenarios.



\subsection{Implementation Details}
We implement $\text{PKS}^4$ based on an CLIP-pretrained ViT backbone~\cite{CLIP}.
We insert the proposed module once after the \textit{8}-th transformer block. 
We uniformly sample $8$ or $32$ frames with a random frame gap, and each frame is resized to $224 \times 224$. 
The hidden dimension of the backbone is $C=512$ (768 for ViT-L), and the state dimension of the selective scanner is set to $d_s=16$. 
In the Kinematic Prior Encoder, we use a temporal neighborhood of 4 frames and a local spatial search window of size $9\times 9$ for correlation modeling.
We train the model using AdamW with a base learning rate of 7e-4 (1e-3 for ViT-L), weight decay of 0.05, and a cosine decay schedule with 5 warm-up epochs. 

\begin{table}[t]
\centering
\resizebox{.84\linewidth}{!}{
\begin{tabular}{lllcc}
\toprule
\multirow{2}{*}{\textbf{Method}} & \textbf{Trainable} & \textbf{Training} & \multirow{2}{*}{\textbf{GFLOPs}} & \textbf{Top-1} \\
& \textbf{Params} & \textbf{Memory} & & (\%)\\
\midrule
\multicolumn{4}{l}{\textit{ViT-B/16}} \\
ST-Adapter~\cite{ST-Adapter} & 7M & 28.8G & 455 & 67.1 \\
AIM~\cite{AIM} & 14M & 35.2G & 624 & 66.4 \\
EVL~\cite{EVL} & 89M & 17.9G & 512 & 61.0 \\
Side4Video~\cite{Side4Video} & 4M & \textbf{8.9G} & 445 & 68.6 \\
PKS$^4$\tiny{(ours)} & 92M & 21.8G & \textbf{437} & \textbf{70.1} \\
\midrule
\multicolumn{4}{l}{\textit{ViT-L/14}} \\
ST-Adapter~\cite{ST-Adapter} & 20M & 51.4G & 2062 & 70.0 \\
AIM~\cite{AIM} & 50M & 64.3G & 2877 & 67.6 \\
EVL~\cite{EVL} & 350M & 33G & 2411 & 65.1 \\
Side4Video~\cite{Side4Video} & 22M & \textbf {15.3G} & 2092 & 70.6 \\
PKS$^4$\tiny{(ours)} & 313M & 20.7G & \textbf{1929} & \textbf{72.9} \\
\bottomrule
\end{tabular}}
\caption{Training memory usage and performance comparison on SSV2.
We report accuracy under the same test setting of $8\times3\times1$ views. 
For ViT-B/16 and ViT-L/14, the batch sizes are 32 and 16, respectively.}
\label{tab:memory_compare_ssv2}
\end{table}
\subsection{Training Cost Comparsion}
\subsubsection{Memory Usage}
We compare PKS$^4$ with existing PEFT methods on SSV2. 
For a fair comparison, we measure memory footprint within the same environment using 8 frames as model input.
All models are evaluated under the same $8\times3\times1$ test views (\#frames × \#clips × \#crops), while the ViT-B/16 and ViT-L/14 variants use batch sizes of 32 and 16, respectively. 
As shown in Table~\ref{tab:memory_compare_ssv2} and Fig~\ref{fig:exp1}(a), PKS$^4$ consistently achieves a better accuracy-memory trade-off than adapter-based baselines. 
Specifically, we also notice that PKS$^4$ with ViT-L/14 does not exhibits significant increase of memory usage than the ViT-B/16 version. 
We suppose this is due to the smaller batch size used for ViT-L/14 (16 \emph{vs.} 32). 
In practice, training memory is dominated more by activation tensors than by parameter count, thus scales strongly with batch size. 
Therefore, the reduced batch size largely compensates for the increased backbone complexity, leading to a lower measured memory footprint for ViT-L/14 in our setting. 
Importantly, this observation also reflects the favorable scaling behavior of PKS$^4$: because its temporal modeling is realized by a single module rather than repeated adapter insertion, its activation memory remains well controlled even on a larger backbone.
\begin{table}[t]
\centering
\resizebox{\linewidth}{!}{
\begin{tabular}{lcccccc}
\toprule
\multirow{2}{*}{\textbf{Method}} & Pre-train & Fine-tune & Pre-train & Fine-tune & Total & Top-1 \\
 & Epochs & Epochs & Compute & Compute & Compute & (\%) \\
\midrule
VideoMamba-M$_{800e}$~\cite{VideoMamba} 
& 800 & 30 
& 27.7883 & 3.7538 & 31.5421 & 70.2 \\
PKS$^4$-B/16\tiny{(ours)} 
& -- & 20 
& -- & 2.9521 & \textbf{2.9521} & 70.3 \\
PKS$^4$-L/14\tiny{(ours)} 
& -- & 15 
& -- & 9.7810 & 9.7810 & \textbf{73.9} \\
\bottomrule
\end{tabular}}
\caption{End-to-end training computation comparison on SSV2.
We report the total computation cost in EFLOPs, including both pre-training and downstream fine-tuning.}
\label{tab:total_compare_ssv2}
\end{table}
\subsubsection{Total Training Computation}
We further compare the total training computation of PKS$^4$ against pure video SSMs. 
As shown in Table~\ref{tab:total_compare_ssv2} and Fig~\ref{fig:exp1}(b), we report the end-to-end compute in EFLOPs, including both video pre-training and downstream fine-tuning.
VideoMamba-M$_{800e}$ relies on large-scale pre-training for 800 epochs, followed by 30 epochs of fine-tuning on SSV2, resulting in a total cost of $31.5421$ EFLOPs. 
Benefiting from preserving spatial structure, PKS$^4$, which was built upon the image-pretrained knowledge, requires only $2.952$ EFLOPs in total, yielding an approximately $10.7\times$ reduction in training computation.
Even the larger PKS$^4$-L/14 variant requires only $9.781$ EFLOPs, which is still about $3.2\times$ lower than VideoMamba-M$_{800e}$.
These results suggest that PKS$^4$ is a more economical route to video understanding. 

\begin{table*}[t]
    \centering
    \resizebox{.92\textwidth}{!}
    {\begin{tabular}{lllcccccc}
         \toprule
         \textbf{Method} & \textbf{Backbone} & \textbf{Pre-train} & \textbf{Resolution} & \textbf{Views} & \textbf{Trainable Params} & \textbf{GFLOPs} & \textbf{Top-1} (\%) & \textbf{Top-5} (\%) \\
         \midrule
         \multicolumn{8}{l}{\textcolor{gray}{\textit{Spatialtemporal Attention}}} \\
         TimeSformer~\cite{TimeSformer} & ViT-B/16 & IN-21K & 224$\times$224 & 8$\times$1$\times$3 & 121.4M & 1770 & 59.5 &  \\
         TimeSformer-L~\cite{TimeSformer} & ViT-B/16 & IN-21K & 224$\times$224 & 96$\times$1$\times$3 & 121.4M & 7140 & 62.3 &  - \\
         ViViT-L/16$\times$2 FE & ViT-L/16$\times$2 FE & IN-21K + K400 & 224$\times$224 & 32$\times$1$\times$3 & 612M & 11940 & 65.9 & 89.9 \\
         Mformer~\cite{Mformer} & ViT-B/16 & IN-21K + K400 & 224$\times$224 & 16$\times$3$\times$1 & 109.1M & 1109 & 66.5 & 90.1 \\
         Mformer~\cite{Mformer} & ViT-L/14 & IN-21K + K400 & 224$\times$224 & 16$\times$3$\times$1 & 381.9M & 2876.4 & 67.1 & 90.6\\
         SwinTransformer~\cite{SwinTransformer} & ViT-B/16 & K400 & 224$\times$224 & 32$\times$1$\times$3 & 88.8M & 963 & 69.6 & 92.7 \\
         MTV~\cite{MTV} & MTV-B & K400 & 224$\times$224 & 16$\times$4$\times$3 & 310M & 11160 & 67.6 & 90.1 \\
         \midrule
         \multicolumn{8}{l}{\textcolor{gray}{\textit{Parameter-Efficient Fine-Tuning (PEFT) with CLIP pre-trained ViT}}} \\
         ST-Adapter~\cite{ST-Adapter} & ViT-B/16 & CLIP & 224$\times$224 & 32$\times$3$\times$1 & 14.1M & 1955 & 69.5 & 92.6 \\
         ST-Adapter~\cite{ST-Adapter} & ViT-L/14 & CLIP & 224$\times$224 & 32$\times$3$\times$1 & - & 8248 & 72.3 & 93.9 \\
         AIM~\cite{AIM} & ViT-B/16 & CLIP & 224$\times$224 & 32$\times$3$\times$1 & 14M & 2496 & 69.1 & 92.3 \\
         AIM~\cite{AIM} & ViT-L/14 & CLIP & 224$\times$224 & 32$\times$3$\times$1 & 50M & 11508 & 70.6 & 92.7 \\
         DUALPATH~\cite{DUALPATH} & ViT-B/16 & CLIP & 224$\times$224 & 32$\times$1$\times$3 & 13M & 716 & 70.3 & 92.9 \\
         DUALPATH~\cite{DUALPATH} & ViT-L/14 & CLIP & 224$\times$224 & 32$\times$1$\times$3 & 33M & 1932 & 71.4 & 93.4 \\
         EVL~\cite{EVL} & ViT-B/16 & CLIP & 224$\times$224 & 32$\times$3$\times$1 & 28.8M & 2824 & 62.4 & -\\
         EVL~\cite{EVL} & ViT-L/14 & CLIP & 224$\times$224 & 32$\times$3$\times$1 & - & 10418 & 66.7 & - \\
         DiST~\cite{DiST} & ViT-B/16 & CLIP & 224$\times$224 & 32$\times$3$\times$1 & - & 1950 & 70.9 & 92.1 \\
         DiST~\cite{DiST} & ViT-L/14 & CLIP & 224$\times$224 & 32$\times$3$\times$1 & - & 8490 & 73.1 & 93.2 \\
         M$^2$-CLIP~\cite{M2CLIP} & ViT-B/16 & CLIP & 224$\times$224 & 32$\times$1$\times$3 & 16M & 2526 & 69.1 & - \\
         \midrule
         \multicolumn{9}{l}{\textcolor{gray}{\textit{Space State Models}}}\\
         VideoMamba~\cite{VideoMamba} & VideoMamba-S & IN-1K & 224$\times$224 & 8$\times$3$\times$2 & 26M & 204 & 66.6 & 90.4 \\
         VideoMamba~\cite{VideoMamba} & VideoMamba-S & IN-1K & 288$\times$288 & 16$\times$3$\times$2 & 26M & 672 & 68.1 & 91.2 \\
         VideoMamba~\cite{VideoMamba} & VideoMamba-M & IN-1K & 224$\times$224 & 8$\times$3$\times$4 & 74M & 1212 & 67.3 & 91.0 \\
         VideoMamba~\cite{VideoMamba} & VideoMamba-M & IN-1K & 288$\times$288 & 16$\times$3$\times$4 & 74M & 3996 & 68.4 & 91.6 \\
         VideoMamba~\cite{VideoMamba} & VideoMamba-M$_{800e}$ & CLIP & 224$\times$224 & 8$\times$3$\times$2 & 74M & 606 & 70.2 & 92.6 \\
         VideoMamba~\cite{VideoMamba} & VideoMamba-M$_{800e}$ & CLIP & 288$\times$288 & 16$\times$3$\times$2 & 74M & 1998 & 71.4 & 92.9 \\
         \midrule
         $\text{PKS}^4$\tiny{(ours)} & ViT-B/16 & CLIP & 224$\times$224 & 8$\times$2$\times$3 & 91.7M & 874 & 70.3 & 92.4 \\
         $\text{PKS}^4$\tiny{(ours)} & ViT-B/16 & CLIP & 224$\times$224 & 32$\times$1$\times$3 & 91.7M & 1745 & 71.7 & 93.3 \\
         $\text{PKS}^4$\tiny{(ours)} & ViT-L/14 & CLIP & 224$\times$224 & 16$\times$1$\times$3 & 312.8M & 3857.4 & 73.9 & 94.0 \\
         $\text{PKS}^4$\tiny{(ours)} & ViT-L/14 & CLIP & 224$\times$224 & 32$\times$1$\times$3 & 312.8M & 7713.6 & \textbf{74.3} & \textbf{94.3} \\
         \bottomrule
    \end{tabular}}
    \caption{
    Comparison with the state-of-the-art on temporal-heavy dataset SSV2.
    Views = \#frames × \#temporal clips × \#spatial crops.
    ``e'' denotes the number of pre-training epochs for VideoMamba.}
    \label{tab:ssv2}
\end{table*}

\begin{table}[t]
  \centering
    \resizebox{.95\linewidth}{!}
    {
    \begin{tabular}{l c c c}
    \toprule
    \textbf{Method} & \textbf{Views} & \textbf{GFLOPs} & \textbf{Top-1} (\%)\\
    \midrule
    TimeSformer-L~\cite{TimeSformer} & 96$\times$1$\times$3 & 7140 & 80.7\\
    ViViT-L/16$\times$2 FE~\cite{Vivim} & 32$\times$1$\times$3 & 11940 & 81.7\\
    Mformer-B/16~\cite{Mformer} & 16$\times$3$\times$10 & 11085 & 79.7\\
    Mformer-L/14~\cite{Mformer} & 16$\times$3$\times$10 & 28764 & 81.1\\
    SwinTransformer-B/16~\cite{SwinTransformer} & 32$\times$4$\times$3 & 3384 & 82.7\\
    SwinTransformer-L/14~\cite{SwinTransformer}  & 32$\times$4$\times$3 & 7248 & 83.1\\
    MTV-B/16~\cite{MTV} & 32$\times$4$\times$3 & 4790.4 & 81.8\\
    ST-Adapter-B/16~\cite{ST-Adapter} & 32$\times$3$\times$1 & 1821 & 82.7\\
    EVL-B/16~\cite{EVL}& 32$\times$3$\times$1 & 2813 & \textbf{84.2} \\
    M$^2$-CLIP-B/16~\cite{M2CLIP} & 32$\times$4$\times$3 & 10104 & 84.1 \\
    VideoMamba-S~\cite{VideoMamba} & 32$\times$4$\times$3 & 1620 & 81.5\\
    VideoMamba-M~\cite{VideoMamba} & 32$\times$4$\times$3 & 4836 & 82.4\\
    VideoMamba-M$_{800e}$~\cite{VideoMamba} & 32$\times$4$\times$3 & 4836 & 83.9\\
    \midrule
    $\text{PKS}^4$-B/16\tiny{(ours)} & 32$\times$1$\times$3 & 1745 & 82.2\\
    $\text{PKS}^4$-L/14\tiny{(ours)} & 32$\times$1$\times$3 & 7713.6 & 84.1\\
    \bottomrule
  \end{tabular}}
  \caption{Performance comparison on K400. 
  Views = \#frames $\times$ \#temporal clips $\times$ \#spatial crops.
  ``e'' denotes the number of pre-training epochs for VideoMamba.}
  \label{tab:k400}
\end{table}

\subsection{Comparison with State-of-the-Art}
We compare our proposed PKS$^{4}$ with representative state-of-the-art methods across three temporal modeling paradigms: spatialtemporal attention, parameter-efficient fine-tuning (PEFT), and state space models (SSMs).
Table~\ref{tab:ssv2} and Table~\ref{tab:k400} show that PKS$^4$ achieves the strongest results on the temporal heavy SSV2 benchmark while remaining highly competitive on K400.
Benefiting from providing suitable prior context for temporal state transitions, PKS$^4$ consistently achieves the best accuracy among the compared methods. 
Under the 32 frames setting, PKS$^4$-L/14 reaches \textbf{74.3\%} Top-1, outperforming the strongest PEFT baseline DiST-L/14 (73.1\%) by \textbf{1.2\%} and VideoMamba-M$_{800e}$ (71.4\%) by \textbf{2.9\%}, while using fewer GFLOPs (7713.6 \emph{vs.} 8490 for DiST-L/14). 
Even with the smaller ViT-B/16 backbone, PKS$^4$ achieves \textbf{71.7\%} Top-1, surpassing DUALPATH-B/16 (70.3\%) by \textbf{1.4\%}, DiST-B/16 (70.9\%) by \textbf{0.8\%}, and VideoMamba-M$_{800e}$ (70.2\%) by \textbf{1.5\%}.
The result therefore directly supports the central premise of our method: temporal modeling benefits from being driven by \emph{motion-aware kinematic priors} rather than generic flattened tokens. 
On K400, PKS$^4$ remains competitive though the advantage becomes less pronounced, which is consistent with the dataset’s weaker reliance on fine-grained temporal cues and stronger dependence on appearance semantics. 
Specifically, PKS$^4$-L/14 achieves \textbf{84.1\%} Top-1, matching M$^2$-CLIP-B/16 and surpassing VideoMamba-M$_{800e}$ (83.9\%), while requiring substantially fewer GFLOPs than several strong baselines. 
Although it does not exceed the best reported results on K400, this behavior is expected that when temporal discrimination is less dominant, the benefit of explicitly modeling kinematic priors is naturally reduced.
Nevertheless, the fact that PKS$^4$ remains competitive with only a single inserted temporal module indicates that \emph{the proposed design does not sacrifice spatial recognition capacity while improving temporal modeling. }
Taken together, the results across SSV2 and K400 suggest that PKS$^4$ is particularly effective in scenarios where accurate recognition depends on preserving spatial structure and capturing motion-aware temporal transitions simultaneously.

\subsection{Ablation Study}
In this section, we ablate PKS$^4$ with different settings.
We use ViT-B/16 backbone and test view of 8$\times$2$\times$3 as default.
The highlighted setting in the table is served as the final setting of PKS$^4$.

\noindent\textbf{Effectiveness of each module.}
Table~\ref{tab:ab_module} evaluates the contribution of each module in PKS$^4$.
Starting from the baseline ViT with temporal average pooling, introducing KS$^4$ already yields a substantial improvement from 44.0\% to 66.1\% Top-1, indicating that the patch-wise temporal scanner is the primary source of performance gain. 
Building on this, equipping KS$^4$ with the variation-based kinematic prior further improves accuracy to 69.1\%, showing that explicitly modeling temporal variation provides a more informative state input than raw features alone. 
Finally, incorporating the correlation operator on top of the variation branch pushes the performance to 70.3\%, yielding an overall gain of 26.3 points over the baseline with only a modest parameter increase of 5.36M. 
These results validate the effectiveness of the full PKS$^4$ design, and suggest that \emph{the two complementary priors consistently improve the performance}.
\begin{table}[t]
    \centering
    \resizebox{\linewidth}{!}
    {\begin{tabular}{l|cc|cc}
    \toprule
    \textbf{Method} & \textbf{Params} & $\mathbf{\Delta}$ & \textbf{Top-1} (\%) & $\mathbf{\Delta}$ \\
    \midrule
    Baseline & 86.34M & - & 44.0 & - \\
    + KS$^4$ & 90.11M & \cellcolor{gray!10}$\uparrow$3.77M & 66.1 &  \cellcolor{gray!10}$\uparrow$ 22.1 \\
    + KS$^4$ + KPE ($\mathcal{O}_\mathrm{var}$) & 90.86M &\cellcolor{gray!30}$\uparrow$ 4.52M & 69.1 & \cellcolor{gray!30}$\uparrow$ 25.1 \\
    \cellcolor{blue!13}+ KS$^4$ + KPE ($\mathcal{O}_\mathrm{var}$ + $\mathcal{O}_\mathrm{corr}$) & 91.70M  & \cellcolor{gray!50}$\uparrow$ 5.36M & 70.3 & \cellcolor{gray!50}$\uparrow$ 26.3 \\
    \bottomrule
    \end{tabular}}
    \caption{The ablation study of PKS$^4$ modules under SSV2.
    ``Baseline'' denotes the pre-trained ViT which only fine-tunes the image encoder and performs temporal average pooling.}
    \label{tab:ab_module}
\end{table}

\noindent\textbf{Effectiveness of insert position of PKS$^4$.}
\begin{table}[t]
    \centering
    \resizebox{.6\linewidth}{!}
    {\begin{tabular}{c|cc}
    \toprule
    \textbf{after} $l$-th \textbf{layer}& \textbf{Top-1} (\%) & \textbf{Top-5} (\%) \\
    \midrule
    5 & 64.69 & 89.06 \\
    6 & 66.32 & 90.69 \\
    7 & 67.92 & 91.43 \\
    \cellcolor{blue!13}8 & \textbf{68.67} & 91.74 \\
    9 & 68.48 & \textbf{91.80} \\
    10 & 68.13 & 91.33 \\
    11 & 67.66 & 91.13 \\
    \bottomrule
    \end{tabular}}
    \caption{Study on the insert position of PKS$^4$ on SSV2.
    We train and test under the setting of 8$\times$1$\times$3 view.}
    \label{tab:ab_position}
\end{table}
We investigate the effect of inserting PKS$^4$ at different depths of the ViT backbone. 
As shown in Table~\ref{tab:ab_position}, the insertion depth has a clear impact on recognition performance, and the better results are achieved when PKS$^4$ is inserted in the middle-to-late stages of the backbone. 
In particular, inserting the module after the 8-th layer yields the best Top-1 accuracy of 68.67\%.
In contrast, earlier insertion leads to sub-optimal performance, \eg inserting PKS$^4$ after the 5-th layer drops Top-1 to 64.69\%. 
This trend suggests that applying PKS$^4$ too early is less favorable, as the intermediate features are still insufficiently mature for reliable kinematic prior extraction and temporal state modeling. 
On the other hand, inserting the module too late also results in a slight degradation, indicating that the remaining layers become insufficient to fully propagate the temporally enhanced representations. 
These results demonstrate that the proposed module is most effective when inserted after the backbone \emph{has formed sufficiently discriminative spatial semantics while still preserving enough subsequent capacity for temporal refinement}.

\noindent\textbf{Scan methods of PKS$^4$.}
\begin{table}[t]
    \centering
    \resizebox{.95\linewidth}{!}
    {\begin{tabular}{c|ccc}
    \toprule
         \textbf{Method} & \textbf{Params} & \textbf{Total Training Compute} & \textbf{Top-1} (\%) \\
         \midrule
         \cellcolor{blue!13}Unidirectional & 91.7M & \textbf{2.9521E} (20 epochs)& 70.3 \\
         Bidirectional & 95.5M & 3.0404E (20 epochs) & \textbf{70.4} \\
         \bottomrule
    \end{tabular}}
    \caption{Study on the scan method of PKS$^4$ on SSV2.
    We report the total fine-tuning computation cost in EFLOPs.}
    \label{tab:ab_scan}
\end{table}
As shown in Table~\ref{tab:ab_scan}, bidirectional scanning brings only a marginal improvement of 0.1\% Top-1 over the unidirectional variant, while introducing higher parameter count and training compute. 
In contrast, the unidirectional design already achieves competitive performance with lower cost, yielding a more favorable efficiency-accuracy trade-off. 
Given the negligible performance gap, we adopt unidirectional scanning as the default setting in PKS$^4$, as it is more computationally economical without sacrificing effectiveness in practice.

\noindent\textbf{The number of context frames in KPE.}
We investigate the effectiveness of different number of context frames used in KPE.
As shown in Table~\ref{tab:ab_context}, introducing temporal context in KPE consistently improves performance over using no context frames. 
The accuracy increases from 67.14\% to 68.67\% when the number of context frames is enlarged, indicating that a moderate temporal neighborhood provides more informative kinematic priors for subsequent scanning. 
However, further increasing the context to 6 frames leads to a performance drop, which suggests that relatively large temporal context may introduce less relevant motion cues and make the extracted priors less discriminative. 
Therefore, we adopt 4 context frames as the default setting, which provides the best trade-off between temporal context richness and prior quality.

\noindent\textbf{Auxiliary CLS temporal module.}
We further explore different designs for the auxiliary temporal modeling route on the CLS token. 
As shown in Table~\ref{tab:ab_cls}, all temporal variants improve over the baseline, confirming that even a lightweight temporal interaction on the global token is beneficial. 
Among them, TokenT1D achieves the best performance, which suggests that applying auxiliary temporal modeling directly on the CLS token can enhance temporal aggregation while introducing only minimal disturbance to the patch-level spatial representations. 
\begin{table}[t]
    \centering
    \resizebox{.7\linewidth}{!}
    {\begin{tabular}{c |cc}
         \toprule
         \textbf{\# Context Frames} & \textbf{Top-1} (\%) & \textbf{Top-5} (\%) \\
         \midrule
         0 & 67.14 & 91.10 \\
         2 & 67.82 & 91.18 \\
         \cellcolor{blue!13}4 & \textbf{68.67} & \textbf{91.74} \\
         6 & 68.06 & 91.47 \\
         \bottomrule
    \end{tabular}}
    \caption{Study on the number of context frames for KPE on SSV2.
    We train and text under the setting of 8$\times$1$\times$3 view.}
    \label{tab:ab_context}
\end{table}

\begin{table}[t]
    \centering
    \resizebox{.6\linewidth}{!}
    {\begin{tabular}{c|cc}
         \toprule
         \textbf{Method} & \textbf{Top-1} (\%) & \textbf{Top-5} (\%) \\
         \midrule
         Baseline & 67.43 & 90.44 \\
         TSM~\cite{TSM} & 69.44 & 91.80 \\
         \cellcolor{blue!13}TokenT1D~\cite{T1D} & \textbf{70.31} & \textbf{92.36} \\
         TokenShift~\cite{TokenShift} & 68.80 & 91.28 \\
         \bottomrule
    \end{tabular}}
    \caption{Study on the temporal modeling method of CLS tokens on SSV2.}
    \label{tab:ab_cls}
\end{table}

%% file: Sections/5-conclusion.tex
\section{Conclusion}
\label{sec:conclusion}
In this paper, we introduced \emph{\textbf{P}arallel \textbf{K}inematic \textbf{S}elective \textbf{S}tate \textbf{S}pace \textbf{S}canners} (PKS$^4$), a highly efficient framework that alleviates the quadratic complexity of dense spatiotemporal attention and the activation memory bottlenecks inherent to multi-layer adapters. 
By extracting motion-aware kinematic priors and integrating them with data-dependent parallel state-space scanning, PKS$^4$ preserves the innate 2D spatial structures of image-pretrained backbones while achieving linear-complexity temporal modeling.
Our extensive evaluations show that PKS$^4$ achieves strong performance across both temporal-heavy and spatial-heavy benchmarks, with particularly clear gains on SSV2 where fine-grained motion reasoning is critical.
More importantly, the results reveal a broader insight: SSMs become substantially more effective for video understanding when their state transitions are driven by motion-aware priors rather than generic flattened tokens. 
PKS$^4$ also demonstrate significant training-efficiency by (i) reducing 10$\times$ total training compute compared with pure video SSMs, and (ii) maintaining a controlled GPU memory footprint compared with multi-adapter methods, while achieving competitive or superior recognition accuracy.
A current limitation is that the gains are less pronounced on appearance-dominant datasets such as K400, and a promising next step is to explore dynamic gating mechanisms, adaptively scanning static video regions to better preserve spatial appearance.